# A Generalization of Gustafson-Kessel Algorithm Using a New Constraint Parameter

**Vasile Patrascu**
Department of Informatics Technology, TAROM Company
Sos. Bucuresti-Ploiesti, K16.5, Bucharest-Otopeni, Romania
e-mail: vpatrascu@tarom.ro, vpatrascu@caramail.com

**Abstract**

In this paper one presents a new fuzzy clustering algorithm based on a dissimilarity function determined by three parameters. This algorithm can be considered a generalization of the Gustafson-Kessel algorithm for fuzzy clustering.

## 1 Introduction

The Gustafson-Kessel [4] algorithm is an important algorithm for fuzzy clustering. Although there has been developed a more efficient algorithm (Gath-Geva [3]), the Gustafson-Kessel algorithm remains the most utilized, because it does not need the utilization of the exponential function [3]. One of the limits of the Gustafson-Kessel is the fact that it supplies unsatisfactory results when clusters having very different volumes are to be separated [5], [6], and [7].

In this paper one presents an algorithm that can eliminate this insufficiency. The algorithm presented in this paper can be considered as a generalization of the Gustafson-Kessel algorithm.

Further the paper has the following structure: section 2 does a short presentation of the Gustafson-Kessel algorithm; section 3 presents the new algorithm; section 4 presents experimental results; section 5 presents some conclusions.

## 2 The Gustafson-Kessel Algorithm

The Gustafson-Kessel algorithm was proposed in [4] as an improvement of the fuzzy C-means clustering algorithm [1, 2]. Let there be the objective function:

$$J(W,M) = \sum_{i=1}^{N} \sum_{j=1}^{c} (w_{ij})^\alpha D_j(x_i) \qquad (1)$$

where $X = \{x_i \in R^k \mid i \in [1,N]\}$ is a set containing $N$ unlabelled vectors; $M = \{m_j \in R^k \mid j \in [1,c]\}$ is the set of centers of clusters; $W = [w_{ij}]$ is the $N \times c$ fuzzy c-partition matrix, containing the membership values of all $x_i$ in all clusters; $D_j(x_i)$ is a dissimilarity measure between the vector $x_i$ and the center $m_j$ of a specific cluster $j$ defined by:

$$D_j(x_i) = \left(\lambda_j \cdot \det(C_j)\right)^{\frac{1}{k}} \cdot d_{ij}^2 \qquad (2)$$

where $C_j$ is the fuzzy covariance matrix of the $j$ cluster defined by:

$$C_j = \frac{\sum_{i=1}^{N} (w_{ij})^\alpha \cdot (x_i - m_j) \cdot (x_i - m_j)^T}{\sum_{i=1}^{N} (w_{ij})^\alpha} \qquad (3)$$

$d_{ij}$ is the Mahalanobis distance

$$d_{ij}^2 = (x_i - m_j)^T \cdot C_j^{-1} \cdot (x_i - m_j) \qquad (4)$$

and $\lambda_1, \lambda_2, ..., \lambda_c$ are $c$ positive constants. Usually $\lambda_j = 1$; $\alpha \in (1, \infty)$ is a control parameter of fuzziness. Usually $\alpha = 2$. The clustering problem can be defined as the minimization of $J$ under the following constraint:

$$\sum_{j=1}^{c} w_{ij} = 1 \qquad (5)$$

The Gustafson-Kessel algorithm consists in the iteration of the following formulae:

$$m_j = \frac{\sum_{i=1}^{N}(w_{ij})^{\alpha} \cdot x_i}{\sum_{i=1}^{N}(w_{ij})^{\alpha}} \qquad (6)$$

and

$$w_{ij} = \frac{1}{\sum_{t=1}^{c}\left(\frac{D_j(x_i)}{D_t(x_i)}\right)^{\frac{1}{\alpha-1}}} \qquad (7)$$

The major drawback is that Gustafson-Kessel algorithm is restricted to constant volume clusters due to the fixed a priori values $\lambda_j$.

## 3 The New Fuzzy Clustering Algorithm

### 3.1 The 3-parameter Dissimilarity Function

Let there be the vector set $X = \{x_1, x_2,...,x_N\}$ where $x_i \in R^k$ and let $c$ be the number of clusters that must be obtained. The cluster $j$ will be described by the parameters: $m_j$ the cluster center, $C_j$ the fuzzy covariance matrix, $f_j$ a parameter that shows how large or how small is the cluster $j$ in comparison to the other clusters, $w_{ij}$ the fuzzy membership degree of the element $x_i$. One denotes:

$$n_j = \sum_{i=1}^{N}(w_{ij})^{\alpha} \qquad (8)$$

$$V_j = \sqrt{\det(C_j)} \qquad (9)$$

$$\rho_j = \frac{n_j}{V_j} \qquad (10)$$

$n_j$ is very close to the cluster cardinality, $V_j$ can be considered as a measure for cluster volume and $\rho_j$ a measure for cluster density. In the framework of this algorithm the dissimilarity function of the element $x_i$ to the cluster $j$ is defined by the relation:

$$D_{ij} = \left(\frac{V_j}{f_j}\right)^{\frac{2}{k}} \cdot d_{ij}^2 \qquad (11)$$

where the distance $d_{ij}$ is defined by relation (4).

In order to solve the problem there will be used the following objective function:

$$J(W,M,F) = \sum_{j=1}^{c}\sum_{i=1}^{N}(w_{ij})^{\alpha} \cdot D_j(x_i) \qquad (12)$$

where $F = \{f_j \in (0,1) \mid j \in [1,c]\}$ and $M$, $W$ have the same definition used by the Gustafson-Kessel algorithm. The parameters $f_j$ verify the following constraint:

$$\sum_{j=1}^{c} f_j = 1 \qquad (13)$$

In addition to this, there are the well-known constraints for the membership degrees $w_{ij}$.

$$\forall i = 1,2,...,N \qquad \sum_{j=1}^{c} w_{ij} = 1 \qquad (14)$$

In this paper there will be presented only the determination of formulae for the parameter $f_j$, due to the fact that the parameters $m_j$ and $w_{ij}$ are computed using formulae that are similar to those used by the Gustafson-Kessel algorithm.

### 3.2 An Equivalent Form for the Objective Function $J$

In this section we will prove the following equivalent form for the objective function $J$.

$$J = \sum_{j=1}^{c}\left(\frac{V_j}{f_j}\right)^{\frac{2}{k}} \cdot k \cdot n_j \qquad (15)$$

This form will be used in the section 3.3. To prove this it is necessary to demonstrate the following relation:

$$\sum_{i=1}^{N}(w_{ij})^{\alpha} \cdot d_{ij}^2 = k \cdot n_j \qquad (16)$$

Let there be $H_j$ the orthogonal matrix that diagonalizes the covariance matrix $C_j$ and $\sigma_{j1}^2, \sigma_{j2}^2, ..., \sigma_{jk}^2$ the eigenvalues of the matrix $C_j$.

There will be:

$$H_j^{-1} = H_j^T \tag{17}$$

$$C_j^{-1} = H_j^T \cdot S_j^{-1} \cdot H_j \tag{18}$$

$$S_j = H_j \cdot C_j \cdot H_j^T \tag{19}$$

where $S_j$ is the following diagonal matrix:

$$S_j = diag(\sigma_{j1}^2, \sigma_{j2}^2, ..., \sigma_{jk}^2) \tag{20}$$

Denoting:

$$y_i = H_j \cdot x_i \tag{21}$$

$$\mu_j = H_j \cdot m_j \tag{22}$$

it results:

$$y_i - \mu_j = H_j \cdot (x_i - m_j) \tag{23}$$

From (4), (18) and (23) it results:

$$d_{ij}^2 = (y_i - \mu_j)^T \cdot S_j^{-1} \cdot (y_i - \mu_j) \tag{24}$$

or

$$d_{ij}^2 = \sum_{t=1}^{k} \frac{(y_{it} - \mu_{jt})^2}{\sigma_{jt}^2} \tag{25}$$

From (19) and (3) it results:

$$S_j = \frac{\sum_{i=1}^{N}(w_{ij})^\alpha \cdot H_j(x_i - m_j)(x_i - m_j)^T H_j^T}{\sum_{i=1}^{N}(w_{ij})^\alpha} \tag{26}$$

Using (23) the relation (26) becomes:

$$S_j = \frac{\sum_{i=1}^{N}(w_{ij})^\alpha \cdot (y_i - \mu_j) \cdot (y_i - \mu_j)^T}{\sum_{i=1}^{N}(w_{ij})^\alpha} \tag{27}$$

From (27) and (20) it results for $t = 1, 2, ..., k$

$$\sigma_{jt}^2 = \frac{\sum_{i=1}^{N}(w_{ij})^\alpha \cdot (y_{it} - \mu_{jt})^2}{\sum_{i=1}^{N}(w_{ij})^\alpha} \tag{28}$$

Using (8) relation (28) becomes:

$$\frac{\sum_{i=1}^{N}(w_{ij})^\alpha \cdot (y_{it} - \mu_{jt})^2}{\sigma_{jt}^2} = n_j \tag{29}$$

From (25) it results:

$$\sum_{i=1}^{N}(w_{ij})^\alpha \cdot d_{ij}^2 = \sum_{t=1}^{k}\left(\sum_{i=1}^{N}\frac{(w_{ij})^\alpha \cdot (y_{it} - \mu_{jt})^2}{\sigma_{jt}^2}\right) \tag{30}$$

Using (29) relation (30) becomes:

$$\sum_{i=1}^{N}(w_{ij})^\alpha \cdot d_{ij}^2 = k \cdot n_j \tag{31}$$

namely the relation (16) that might be demonstrated.

### 3.3 The Calculus of the Parameters $f$

In this section it is presented the determination of the parameters $f_j$. There will be considered the objective function:

$$J = \sum_{j=1}^{c}\left(\frac{V_j}{f_j}\right)^{\frac{2}{k}} \cdot k \cdot n_j + \omega\left(\sum_{j=1}^{c} f_j - 1\right) \tag{32}$$

where $\omega$ is the Lagrange multiplier.

We must solve the equation:

$$\frac{\partial J}{\partial f_j} = 0 \tag{33}$$

From (32) and (33) it results:

$$-\frac{2}{k}\frac{1}{f_j^{1+\frac{2}{k}}} \cdot n_j \cdot k \cdot V_j^{\frac{2}{k}} + \omega = 0 \tag{34}$$

From (34) it results:

$$f_j = \left(\frac{2}{\omega}\right)^{\frac{k}{k+2}} \cdot \left(n_j^k \cdot V_j^2\right)^{\frac{1}{k+2}} \tag{35}$$

Taking into account the relations (13) and (35) it results:

$$\left(\frac{2}{\omega}\right)^{\frac{k}{k+2}} = \frac{1}{\sum_{t=1}^{c} \left(n_t^k \cdot V_t^2\right)^{\frac{1}{k+2}}} \quad (36)$$

and

$$f_j = \frac{\left(n_j^k \cdot V_j^2\right)^{\frac{1}{k+2}}}{\sum_{t=1}^{c} \left(n_t^k \cdot V_t^2\right)^{\frac{1}{k+2}}} \quad (37)$$

From (10) and (37) it results the following equivalent formula for the parameter $f_j$:

$$f_j = \frac{\rho_j^{\frac{k}{k+2}} \cdot V_j}{\sum_{t=1}^{c} \rho_t^{\frac{k}{k+2}} \cdot V_t} \quad (38)$$

### 3.4 The Calculus of the Parameters $m$

The calculus formulae for the cluster centers $m_j$ are similar to those used by the Gustafson-Kessel algorithm, namely:

$$m_j = \frac{\sum_{i=1}^{N} (w_{ij})^\alpha \cdot x_i}{\sum_{i=1}^{N} (w_{ij})^\alpha} \quad (39)$$

### 3.5 The Calculus of the Parameters $w$

The calculus formulae for the membership degrees $w_{ij}$ are similar to those used by the Gustafson-Kessel algorithm, namely:

$$w_{ij} = \frac{1}{\sum_{t=1}^{c} \left(\frac{D_j(x_i)}{D_t(x_i)}\right)^{\frac{1}{\alpha-1}}} \quad (40)$$

But from (38) it results:

$$\frac{V_j}{f_j} = \sum_{t=1}^{c} \left(\frac{\rho_t}{\rho_j}\right)^{\frac{k}{k+2}} \cdot V_t \quad (41)$$

Using (41) relation (11) becomes:

$$D_{ij} = \left(\sum_{t=1}^{c} \left(\frac{\rho_t}{\rho_j}\right)^{\frac{k}{k+2}} \cdot V_t\right)^{\frac{2}{k}} \cdot d_{ij}^2 \quad (42)$$

From (40) and (42) one can obtain the following equivalent formulae for $w_{ij}$.

$$w_{ij} = \frac{\left(\rho_j^{\frac{2}{k+2}} \cdot \frac{1}{d_{ij}^2}\right)^{\frac{1}{\alpha-1}}}{\sum_{t=1}^{c} \left(\rho_t^{\frac{2}{k+2}} \cdot \frac{1}{d_{it}^2}\right)^{\frac{1}{\alpha-1}}} \quad (43)$$

## 4 Experimental Results

One presents the using of the new algorithm for the two test sets shown in the figures 1 and 5. The first set was obtained by a random generation of some points inside two ellipses (figure 1). The second set was obtained by a random generating of some points inside three ellipses (figure 5). The obtained results using the new algorithm can be seen in the figures 3 and 7 respectively, those obtained using the Gustafson-Kessel algorithm in the figures 2 and 6, and those obtained using Gath-Geva algorithm in figures 4 and 8. One can see that while the Gustafson-Kessel algorithm has generated clusters having approximately the same size ( figures 2 and 6), the new algorithm (figures 3 and 7) has generated clusters that are very close to the ideal clustering (figures 1 and 5). The results obtained using the proposed algorithm are very close to those obtained using the Gath-Geva algorithm.

## 5 Conclusions

This paper has presented an algorithm that can be considered a generalization of Gustafson-Kessel algorithm. There has been used a dissimilarity function having three parameters and a new constraint. In this way there it was eliminated one of the main insufficiency of the Gustafson-Kessel algorithm regarding to the obtaining of some clusters having very different sizes. The experimental results stand as proof of the new algorithm superiority in comparison to the Gustafson-Kessel one.

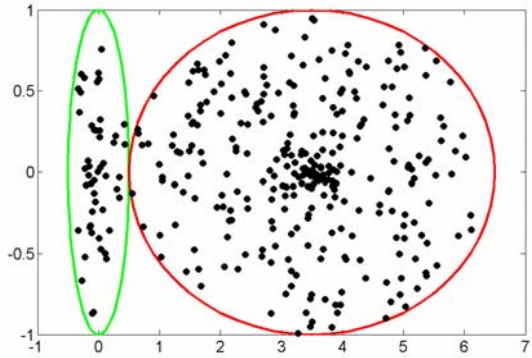

Figure 1. The test set (I).

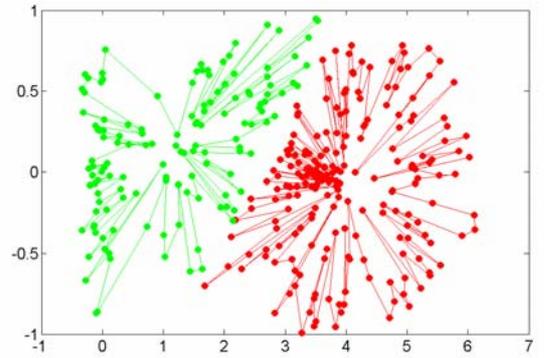

Figure 2. The Gustafson-Kessel clustering

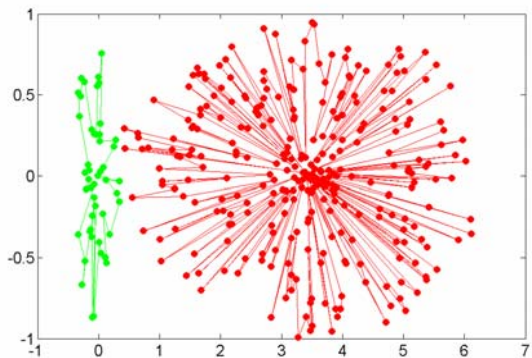

Figure 3. The generalized Gustafson-Kessel clustering.

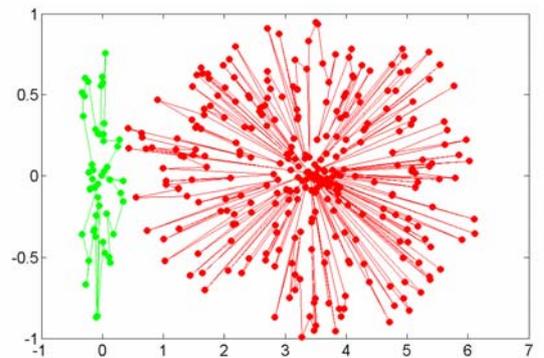

Figure 4. The Gath-Geva clustering

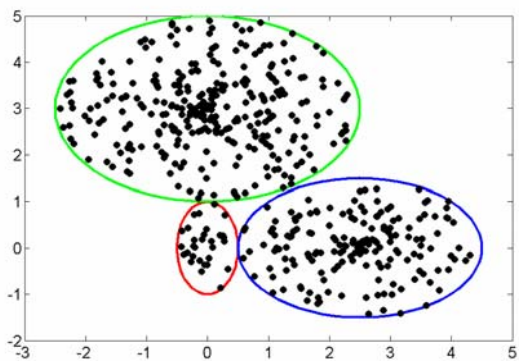

Figure 5. The test set (II).

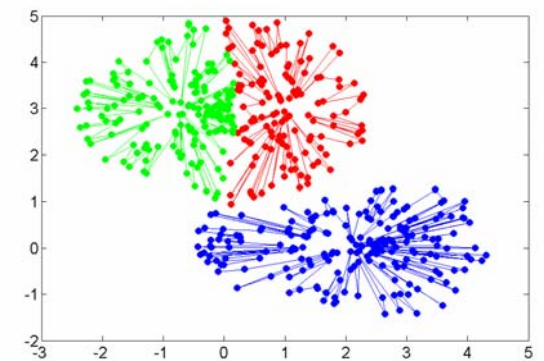

Figure 6. The Gustafson-Kessel clustering

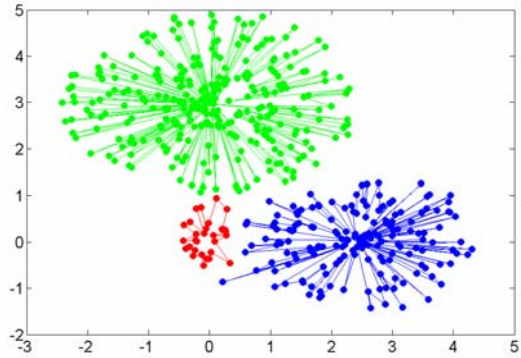

Figure 7. The generalized Gustafson-Kessel clustering.

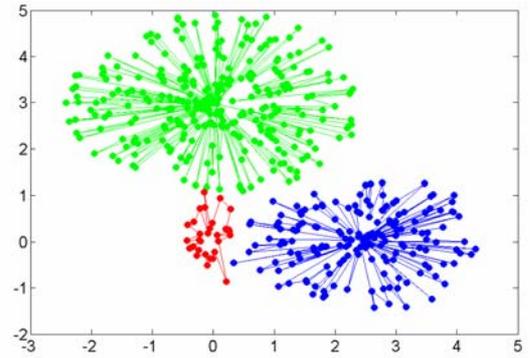

Figure 8. The Gath-Geva clustering.

## References

[1] J. C. Bezdek, *Fuzzy mathematics in pattern classification,* Ph.D. dissertation, Cornell Univ., Ithaca, NY, 1973.

[2] J. C. Bezdek, *Pattern recognition with fuzzy objective function algorithms*, New York: Plenum Press, 1981.

[3] I. I. Gath and A. B. Geva, "Unsupervised optimal fuzzy clustering", *IEEE Trans. Pattern And Machine Intell.*, Vol. 2, no. 7, pp. 773-781, 1989.

[4] E. E. Gustafson and W. C. Kessel, "Fuzzy clustering with a fuzzy covariance matrix", *Proc. IEEE CDC*, San-Diego, CA, pp. 761-766, 1979.

[5] R. A. Keller and F. Klawonn, "Adaptation of cluster sizes in objective function based fuzzy clustering", In T.Leondes, editor. Intelligent Systems: Techniques and Applications – Database and Learning Systems, vol. 4, pp. 181-191. CRC Press, 2002.

[6] R. Krishnapuram and J. Kim, "Clustering algorithms based on volume criteria". *IEEE Transactions on Fuzzy Systems*, 8(2), pp. 228-236, 2000.

[7] M. Setnes and U. Kaymak, "Extended fuzzy c-means with volume prototypes and cluster merging", *In Proc. Of 6[th] European Congress on Intelligent Techniques and Soft Computing*, pp. 1360-1364, 1998.